\documentclass[10pt,twocolumn,letterpaper]{article}

\usepackage{iccv}
\usepackage{times}
\usepackage{epsfig}
\usepackage{graphicx}
\usepackage{epstopdf}
\usepackage{amsmath}
\usepackage{amssymb}

\usepackage{enumitem}
\usepackage{subfigure}
\usepackage{multirow}
\usepackage{algorithm}
\usepackage{algorithmicx}
\usepackage{algcompatible}
\usepackage{float}
\usepackage[flushleft]{threeparttable}

\newcommand{\tabincell}[2]{\begin{tabular}{@{}#1@{}}#2\end{tabular}}

\hypersetup{breaklinks=true,bookmarks=false,colorlinks=true,urlcolor=blue}

\iccvfinalcopy 

\def\iccvPaperID{****} 

\ificcvfinal\fi

\begin{document}

\title{Differentiable Soft Quantization:\\
Bridging Full-Precision and Low-Bit Neural Networks}

\author{
Ruihao Gong$^{1, 2}$ \quad
Xianglong Liu$^1$\thanks{corresponding author}\quad
Shenghu Jiang$^{1, 2}$ \quad
Tianxiang Li$^{2, 3}$\\
Peng Hu$^{2}$ \quad
Jiazhen Lin $^{2}$ \quad
Fengwei Yu$^2$ \quad
Junjie Yan$^2$\\
$^1$State Key Laboratory of Software Development Environment, Beihang University\\
$^2$SenseTime Group Limited \qquad
$^3$Beijing Institute of Technology\\
{\tt\small \{gongruihao, xlliu, jiangshenghu\}@nlsde.buaa.edu.cn}\\
{\tt\small \{litianxiang, hupeng, linjiazhen, yufengwei, yanjunjie\}@sensetime.com}
}


\maketitle
\ificcvfinal\thispagestyle{empty}\fi

\begin{abstract}
    Hardware-friendly network quantization (e.g., binary/uniform quantization) can efficiently accelerate the inference and meanwhile reduce memory consumption of the deep neural networks, which is crucial for model deployment on resource-limited devices like mobile phones. However, due to the discreteness of low-bit quantization, existing quantization methods often face the unstable training process and severe performance degradation. To address this problem, in this paper we propose Differentiable Soft Quantization (DSQ) to bridge the gap between the full-precision and low-bit networks. DSQ can automatically evolve during training to gradually approximate the standard quantization. Owing to its differentiable property, DSQ can help pursue the accurate gradients in backward propagation, and reduce the quantization loss in forward process with an appropriate clipping range. Extensive experiments over several popular network structures show that training low-bit neural networks with DSQ can consistently outperform state-of-the-art quantization methods. Besides, our first efficient implementation for deploying 2 to 4-bit DSQ on devices with ARM architecture achieves up to 1.7$\times$ speed up, compared with the open-source 8-bit high-performance inference framework NCNN \cite{ncnn}.
\end{abstract}

\section{Introduction}
Deep convolution neural networks have achieved great success in many fields, such as computer vision, natural language processing, information retrieval, etc. However, the expensive memory and computation costs seriously impede their deployment on the widespread resource-limited devices, especially for real-time applications. To address this problem, the quantization technique has emerged as a promising network compression solution and achieved substantial progress in recent years. It can largely reduce the network storage and meanwhile accelerate the inference speed using different types of quantizers, mainly including binary/ternary \cite{BinaryConnect,BNN+-1,BNN,TWN,TTQ,ABCNet,XnorNet}, uniform \cite{AdaptiveLQ,Apprentice,BridgingGap,DiscoveringLowPrecision,Dorefa,GoogleCVPR2018,HAQ,INQ,LearnInterval,PACT,PostTraining4bit,ReLeQ,Whitepaper} and non-uniform \cite{BalancedQ,ClipQ,DeepCompression,HWGQ,LQNet,LogQ,SLQ,TSQ,UNIQ}.

Limited by the specific hardware features like the instruction sets, most quantization methods can hardly accomplish the goal of network acceleration and may still heavily depend on the special hardware design and long-term hardware development. For example, a special inference engine EIE \cite{EIE} has been developed to speed up the method in \cite{DeepCompression}. Fortunately, the recent studies have proved that both the binary and uniform quantization models enjoy the hardware-friendly property \cite{GoogleCVPR2018,Gemmlowp,IntelCaffe,TensorRT}, which enables us to accelerate the inference directly on off-the-shelf hardware with the efficient bit operation or integer-only arithmetic. 

Despite the attractive benefits, when quantizing into extremely low bit, existing binary and uniform quantization models still face the severe performance degradation, due to the limited and discrete quantization levels \cite{BridgingGap}. First, based on the discrete quantized representation, the backward propagation can hardly access the accurate gradients, and thus has to resort to the appropriate approximation. In the literature, straight through estimation (STE) \cite{STE} has been widely used for approximation. But it ignores the influence of quantization, and when the data is quantized to extremely low bit, its error will be amplified, causing an obvious instability of optimization. Experiments and analysis in \cite{DiscoveringLowPrecision,BinaryConnect,DeeperUnderstanding} have shown that the gradient error caused by quantization and STE greatly harms the accuracy of quantized models.

Besides, the quantization itself inevitably brings large deviations between the original data and their quantization values, and thus often causes the performance decrease. In practice, the quantization is usually completed by two operations: clipping and rounding. The former confines data to a smaller range, while the latter maps the original value to its nearest quantization point. Both operations contribute to the quantization loss. Therefore, to alleviate the performance degradation, it is also important to find an appropriate clipping range and make a balance between clipping and rounding  \cite{PACT,LearnInterval}.

To solve the problems, in this paper we introduce \textbf{Differentiable Soft Quantization (DSQ)} to well approximate the standard binary and uniform quantization process (see the framework in Figure \ref{fig:overview}). DSQ employs a series of hyperbolic tangent functions to gradually approach the staircase function for low-bit quantization (e.g., sign for 1-bit case), and meanwhile keeps the smoothness for easy gradient calculation. We reformulate the DSQ function with respect to an approximation characteristic variable, and correspondingly develop an evolution training strategy to progressively learn the differential quantization function. During training, the approximation between DSQ and standard quantization can be controlled by the characteristic variable, which together with the clipping values can be automatically determined in the network. Our DSQ decreases deviations caused by extremely low-bit quantization, and thus makes the forward and backward process more consistent and stable in the training.

The specific design makes the DSQ own the following advantages compared to the state-of-the-art solutions:

\begin{figure*}[htbp]
	\begin{center}
		\includegraphics[width=1\linewidth]{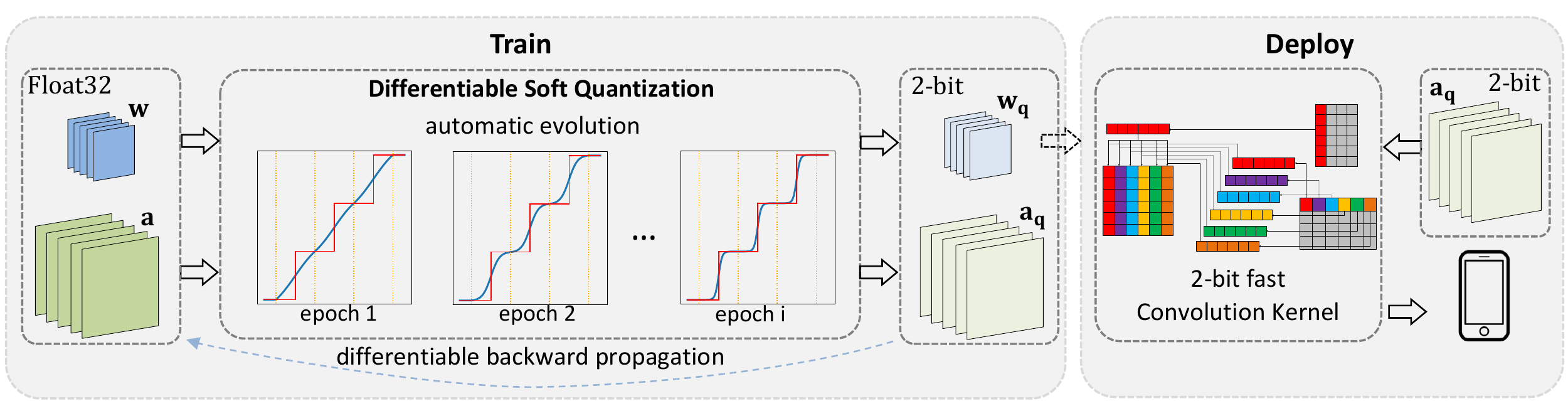}
	\end{center}
	\vspace*{-0.1in}
	\caption{An overview of Differentiable Soft Quantization (DSQ), taking 2-bit uniform quantization as an example. During training, we apply piecewise DSQ to redistribute the data and make it automatically evolve in each epoch to behave more like uniform quantization. After training, the piecewise DSQ can completely convert to the hard uniform quantization by sign operation, ensuring an easy and efficient deployment on resource limited devices.}
	\label{fig:overview}
	\vspace{-0.15in}
\end{figure*} 

\begin{itemize}[leftmargin=*]
    \item \textbf{Novel quantization}. We introduce a DSQ function to well approximate the standard binary and uniform quantization. The approximation of DSQ can be easily controlled in the evolution training way. 
    \item \textbf{Easy convergence}. DSQ acts as a rectifier to gradually redistribute the data according to quantization points. Thus the backward propagation becomes more consistent with the forward pass, leading to an easier convergence with the accurate gradients.
    \item \textbf{Balanced loss}. With the help of DSQ, we can jointly determine the clipping range and approximation of the quantization, and thus balance the quantization loss including clipping error and rounding error.
    \item \textbf{High efficiency}. DSQ can be implemented based on our fast computation kernels, and its inference speed surpasses most open-source high performance inference frameworks.
    \item \textbf{Strong flexibility}. DSQ is compatible with the binary or uniform quantization methods, easy to deploy in state-of-the-art network structures and able to get further accuracy improvement. 
\end{itemize}

\section{Related Work}
\subsection{Network Quantization}
Network quantization aims to obtain low-precision networks with high accuracy. One way to speed up low-precision networks is to utilize bit operation \cite{BNN,BNN+-1,BinaryConnect,ABCNet,TTQ,TWN,DeepCompression,XnorNet}. They quantize weights or activations to \{-1, +1\} or \{-1, 0, +1\}. Because only two or three values can be used, training a binary or ternary model with high accuracy is very challenging. Another way to achieve acceleration is to uniformly convert weight and activations to fix-point representation. \cite{GoogleCVPR2018} has verified the feasibility of 8-bit uniform fix-point quantization. But lower bit quantization faces more challenges on accuracy. To address this problem, \cite{PACT,LearnInterval} try to optimize the clipping value or quantization interval for the task specific loss in an end-to-end manner. \cite{HAQ,AdaptiveLQ,ReLeQ} apply different techniques to find optimal bit-widths for each layer. \cite{INQ} and \cite{LearnInterval} optimize the training process with incremental and progressive quantization. \cite{WRPN} and \cite{BNN} adjust the network structure to adapt to quantization. \cite{Apprentice} introduces knowledge distillation to improve quantized networks' performance. \cite{LQNet} learns a more flexibility quantizer with basis vector. Besides, there are other non-uniform methods like \cite{LogQ,DeepCompression,UNIQ,ClipQ,TSQ,SLQ,BalancedQ} which may need delicate hardware to get acceleration.

\subsection{Efficient Deployment}
With hardware-friendly network quantization, many efficient deployment frameworks are emerging. NVIDIA TensorRT \cite{TensorRT} is a high-performance deep learning inference platform. It provides INT8 optimizations for deployments on GPU devices. Intel Caffe \cite{IntelCaffe} is a fork of official Caffe \cite{caffe} to improve performance on CPU, in particular Intel Xeon processors. Gemmlowp is a low-precision GEMM library in tensorflow \cite{TensorFlow} supporting ARM and Intel X86. NCNN \cite{ncnn} is Tencent's inference framework optimized for mobile platforms. These frameworks usually support 8-bit integer arithmetic, but does not do specific optimization for lower bit computation. To validate the efficiency of our quantization method, in this paper we implement 2-bit fast integer arithmetic with ARM NEON technology, which is an advanced SIMD architecture extension for the ARM Cortex-A series and Cortex-R52 processors.

\section{Differentiable Soft Quantization}
In this paper, we consider the standard 1-bit binary and multi-bit uniform quantization.
\subsection{Preliminaries}
For 1-bit binary quantization, the binary neural network (BNN) limits its activations and weights to either -1 or +1, usually using the binary function:
\begin{equation}\label{binary}
Q_B(x) = \text{sgn}(x)=\begin{cases}
+1, &\text{if}\ x\ge 0,\\
-1, &\text{otherwise}.
\end{cases}
\end{equation}
\vspace{-0.02in}
For multi-bit uniform quantization, given the bit width $b$ and the floating-point activation/weight $x$ following in the range $(l, u)$,
the complete quantization-dequantization process of uniform quantization can be defined as:
\vspace{-0.02in}
\begin{equation}
	Q_U(x) =  \text{round}(\frac{x}{\Delta}) \Delta ,
\end{equation}
\vspace{-0.02in}
where the original range $(l, u)$ is divided into $2^b-1$ intervals $\mathcal{P}_i, i \in (0, 1, \ldots, 2^b-1)$, and $\Delta = \frac{u-l}{2^b-1}$ is the interval length.

\subsection{Quantization function}
The derivative of binary/uniform quantization function is zero almost everywhere, which not only makes the training of the quantized network unstable, but also largely decreases the accuracy. To reduce the gap between the full-precision model and its quantized low-precision model, a differentiable asymptotic function is first introduced to approximately model the binary/uniform quantizer. This function handles the point $x$ falling in different intervals $\mathcal{P}_i$:
\begin{equation}\label{DSQi}
	\varphi(x) = s \tanh\left(k(x-m_i)\right),\quad \text{if }x \in \mathcal{P}_i,
\end{equation}
with
\vspace{-0.15in}
\begin{equation}
m_i = l +  (i + 0.5)\Delta\ \text{and}\ s = \frac{1}{\tanh(0.5k\Delta)}.
\end{equation}

\begin{figure}[htbp]
\centering
{\subfigcapskip = -8pt
\subfigure[Piecewise DSQ]{
\begin{minipage}[t]{0.22\textwidth}
\centering
\includegraphics[width=1\linewidth]{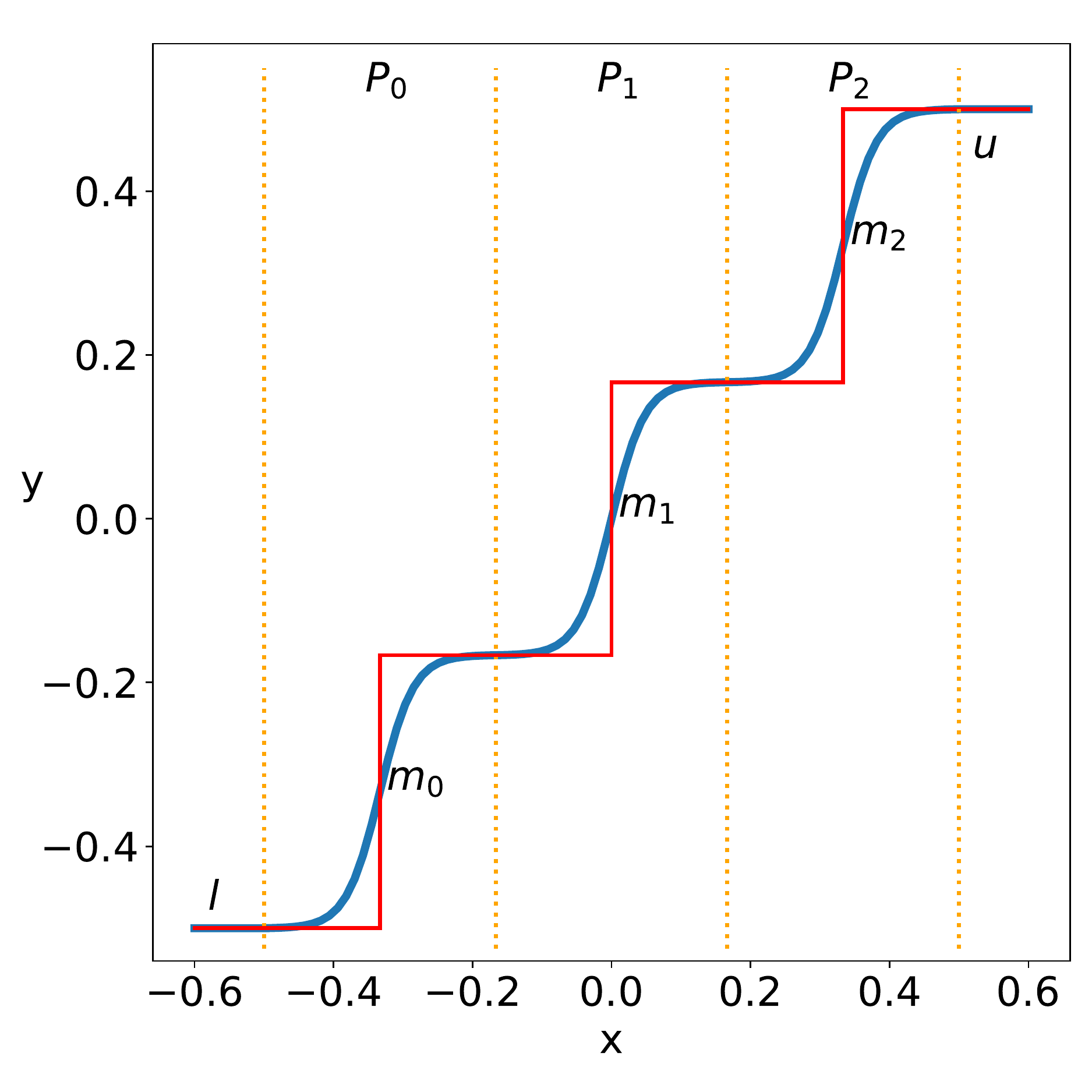}
\end{minipage}
}}
{\subfigcapskip = -8pt
\subfigure[Binary DSQ]{
\begin{minipage}[t]{0.22\textwidth}
\centering
\includegraphics[width=1\linewidth]{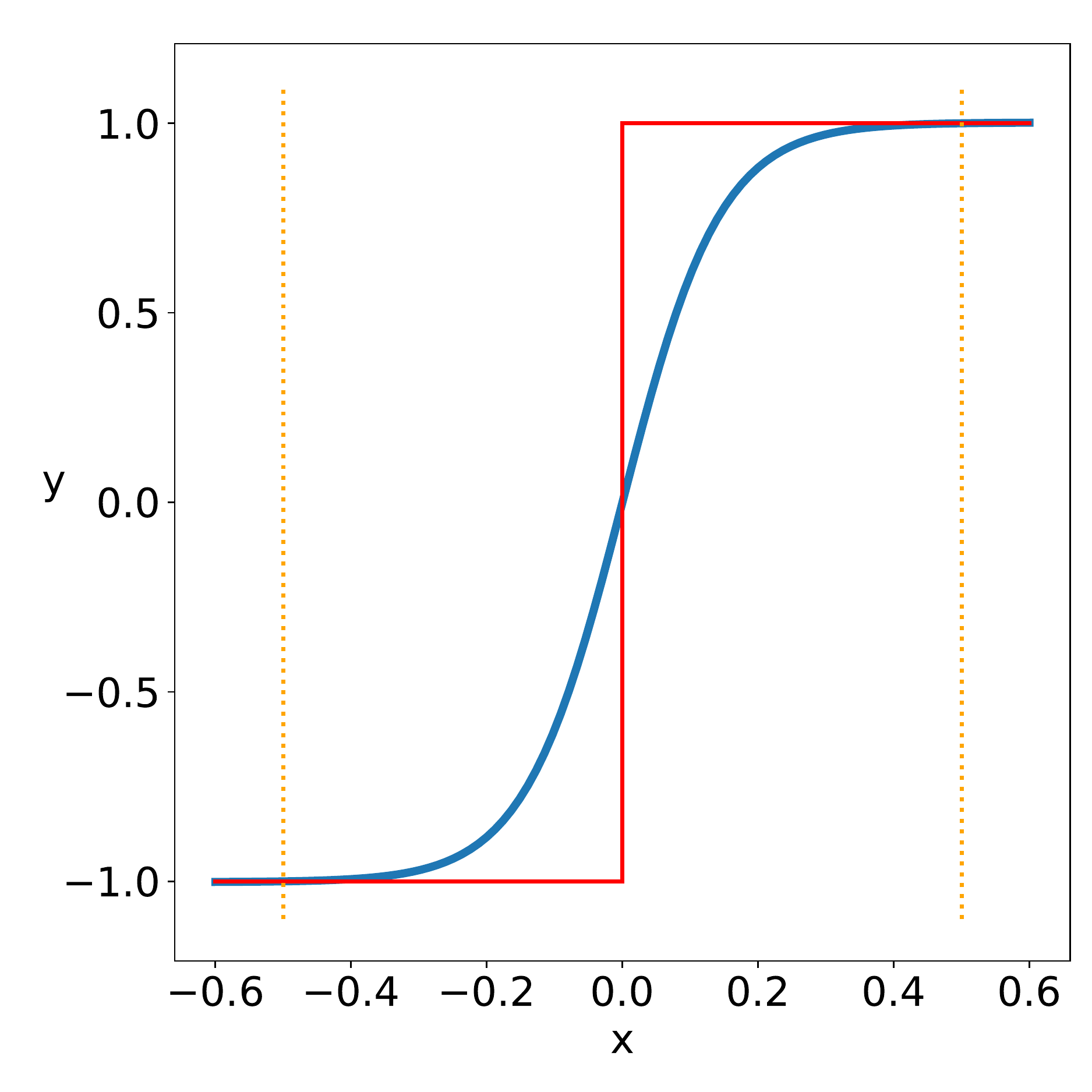}
\end{minipage}
}}
\caption{Curves of differentiable soft quantization function for (a) uniform quantization, (b) binary quantization, a special case of (a).}\label{dsqfunc}
\vspace{-0.15in}
\end{figure}

The scaling parameter $s$ guarantees that $\tanh$ functions of $\varphi$ for the adjacent intervals can be smoothly connected (see Figure \ref{dsqfunc}(a)). Owing to the highly symmetry of the $\tanh$ function,  $\varphi$ will be continuously differentiable everywhere. Besides, the coefficient $k$ determines the shape of the asymptotic function. Namely, the larger $k$ is, the more the asymptotic function behaves like the desired staircase function generated by uniform quantizer with multiple quantization levels. 

Based on the asymptotic function $\varphi$, we can have our differentiable soft quantization (DSQ) function, approximating the uniform quantizer:
\begin{align}
{Q}_{S}(x)=\begin{cases}
\qquad \qquad l, &x<l,\\
\qquad \qquad u, &x>u,\\
l+\Delta\left(i+\frac{\varphi(x)+1}{2}\right), &x \in \mathcal{P}_i
\end{cases}
\end{align}

As shown in Figure \ref{fig:overview}, the curve of the asymptotic function gradually approaches the piecewise curve of the uniform quantizer, when $k$ becomes larger. Thus in practice we can also utilize it to simulate the influence of real quantization in the forward pass and this function is well behaved for purposes of calculating gradients in the backward pass. From another view, DSQ acts as a rectifier, which align the data with its quantization points with small quantization error simply by the redistribution. Subsequently, during the backward propagation, the gradient can better reflect the correct direction of updating.

When $\varphi$ composites with the sign function, DSQ can serve as a piecewise uniform quantizer (see Figure \ref{dsqfunc}(a)). It is worth mentioning that when there is only one interval, we can also simulate the standard model binarization in this way (see Figure \ref{dsqfunc}(b)). This means the binary quantization in (\ref{binary}) can also be treated as a special case of our soft quantization function.

\subsection{Evolution to the standard quantization}
With the DSQ function, we can easily find a differentiable substitute for standard quantization. However, how well the DSQ approximates the standard quantization can largely affect the behaviors of the quantized models. Namely, it is highly required we can adaptively choose the appropriate parameters of DSQ in the training process, and thus promise the controlled approximation according to the optimization goal of the quantized network.

\begin{figure}[htbp]
\centering
{\subfigcapskip = -8pt
\subfigure[Characteristic of $\alpha$]{
\begin{minipage}[t]{0.22\textwidth}
\centering
\includegraphics[width=1\linewidth]{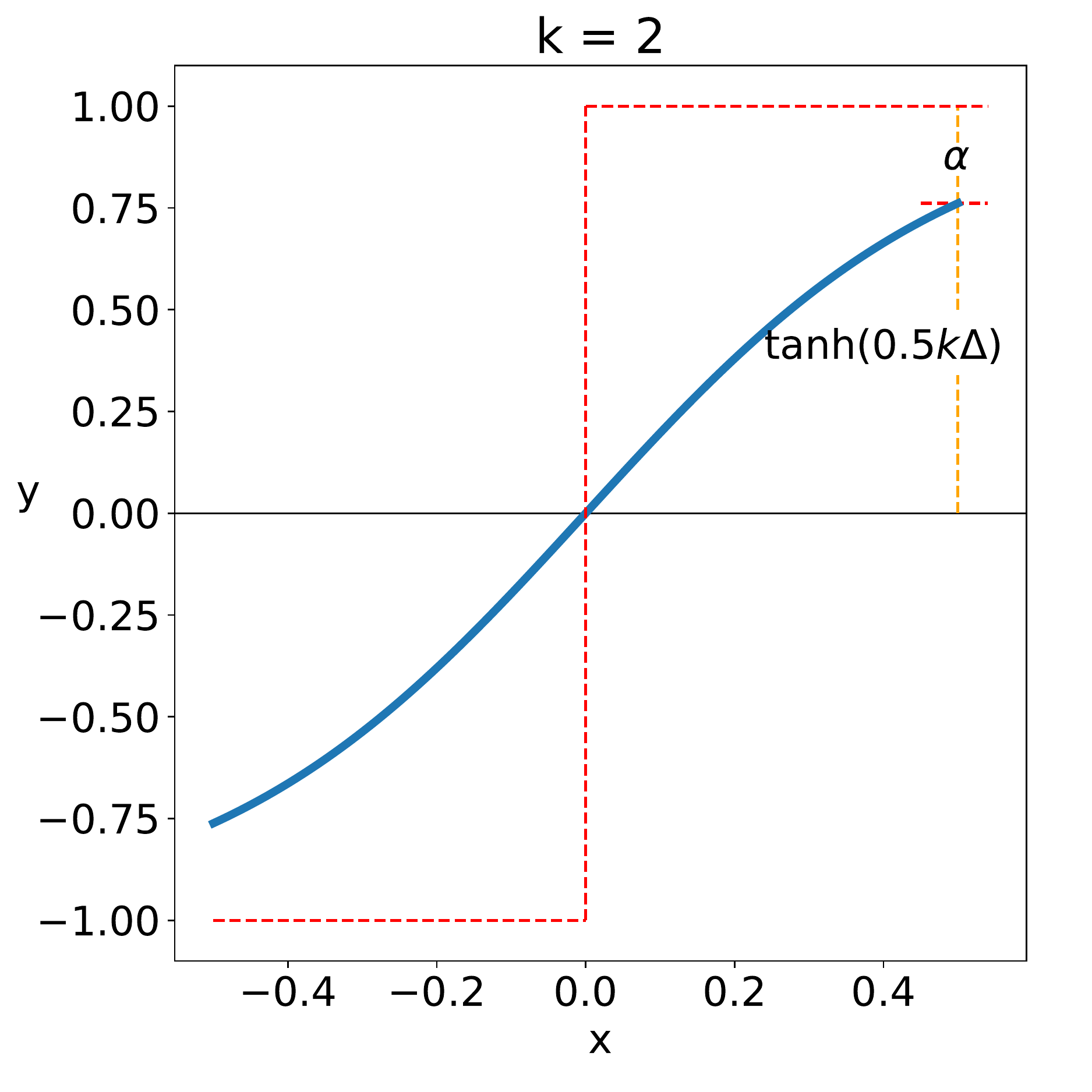}
\end{minipage}
}}
{\subfigcapskip = -8pt
\subfigure[Effect of $\alpha$]{
\begin{minipage}[t]{0.22\textwidth}
\centering
\includegraphics[width=1\linewidth]{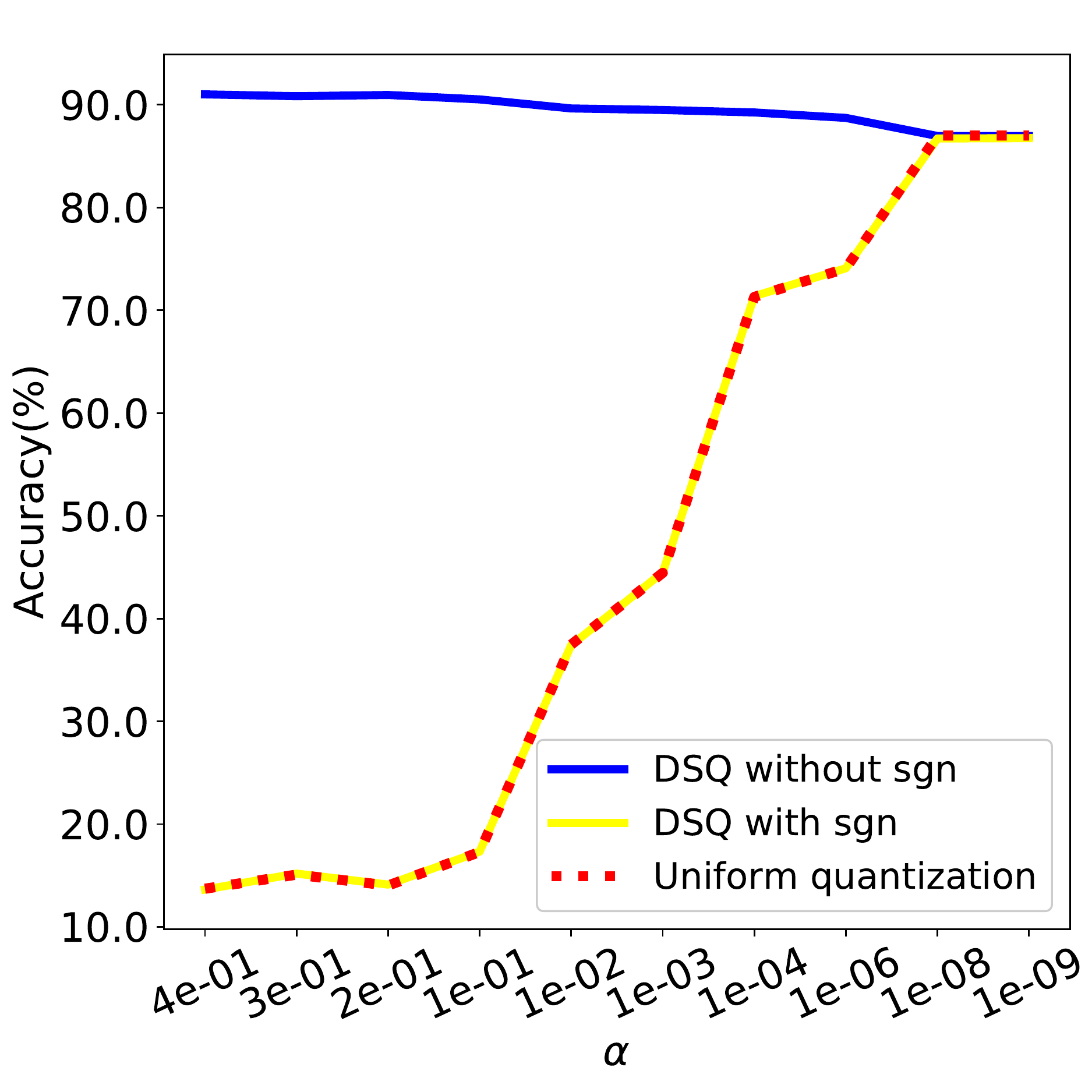}
\end{minipage}
}}
\caption{The characteristic variable $\alpha$ in DSQ.}\label{fig:value_metric}
\vspace{-0.08in}
\end{figure}

To achieve this goal, it is important to find a characteristic variable to measure the approximation between DSQ and standard quantization. Figure \ref{fig:value_metric}(a) shows the curve of DSQ in one interval without scaling to $-1$ and $+1$. It is easy to prove that when the distance from the maximal point in the curve to the upper bound $+1$ is small enough, DSQ function can perfectly approximates the standard quantizer. Based on this observation, we introduce the characteristic variable $\alpha$ as follows:
\vspace{-0.05in}
\begin{equation}
\label{alpha_value}
\alpha = {1-\tanh(0.5 k \Delta)}=1-\frac{1}{s}.
\end{equation}
\vspace{-0.35in}

Furthermore, we can reformulate the DSQ function only with respect to the parameter $\alpha$ and $\Delta$. Specifically, based on the above equation, we have 
\vspace{-0.05in}
\begin{equation}
s=\frac{1}{1-\alpha}.
\end{equation}
With the facts $\Delta=\frac{u-l}{2^b-1}$ and $\varphi(0.5\Delta)=1$, we also have 
\begin{equation}
\label{value_metric_equation}
k =\frac{1}{\Delta} \log(\frac{2}{\alpha}-1).
\end{equation}

Figure \ref{fig:value_metric}(b) illustrates the effect of $\alpha$, where 2-bit quantized models are first trained using DSQ with respect to different $\alpha$. The curves respectively show the inference accuracy performance of DSQ, DSQ appended with sign function, and standard uniform quantization. It is easy to see that our DSQ with sign function can perform quite consistently with the uniform quantization. Especially, when $\alpha$ is small, DSQ can well approximate the performance of the uniform quantization. This means an appropriate $\alpha$ will enble DSQ to help train a quantized model with high accuracy. 

Now we can see that the approximation power of DSQ actually depends on the appropriate $\alpha$, which plays an important role in the optimization. To adaptively determine it, we introduce an evolution training strategy, which takes $\alpha$ as a variable to be optimized in the quantized network. In this way, we can adaptively adjust $\alpha$ and force DSQ to evolve into the standard quantizer during the training. Formally, we can formulate it as the network loss minimization problem with respect to the input $x$ and output $y$ at each layer:
\vspace{-0.05in}
\begin{equation}\label{loss_function}
\min\limits_{\alpha} \mathcal{L}(\alpha; x, y) \quad \text{ s.t. } ||\alpha||_2 < \lambda
\end{equation}
According to the formulation, we can calculate the gradient of $\alpha$ in the backward process and then automatically adjust it with the L2 regularization constraint.
\vspace{-0.05in}
\begin{equation}
\label{alpha_derivate}
\frac{\partial{y}}{\partial{\alpha}}=\begin{cases}
0, &x< l,\\
0, &x>u,\\
\frac{\partial{Q_{S}(x)}}{\partial{\alpha}}, &x \in P_i\\
\end{cases}
\end{equation}

\subsection{Balancing clipping error and rounding error}
Clipping and rounding together cause the quantization error. Usually, when the quantizer clips more, the clipping error increases and the rounding error decreases. Owing to the differentiable evolution from the soft quantizer to the standard quantizer, we can further analyze the connections between clipping error and rounding error in DSQ. Specifically, we can jointly optimize the lower bound and upper bound of clipping to pursue a balance between clipping error and rounding error. 
\begin{align}
\label{d_l}
    &\frac{\partial{y}}{\partial{l}}=\begin{cases}
	\qquad \qquad 1, &x<l,\\
	\qquad \qquad 0, &x>u,\\
	1+q\frac{\partial{\Delta}}{\partial{l}}+\frac{\Delta}{2}
	\frac{\partial{\varphi(x)}}{\partial{l}}, &x \in P_i\\
	\end{cases}\\
\label{d_u}
	&\frac{\partial{y}}{\partial{u}}=\begin{cases}
\qquad \qquad 0, &x<l,\\
\qquad \qquad 1, &x>u,\\
q\frac{\partial{\Delta}}{\partial{u}}+\frac{\Delta}{2}
    \frac{\partial{\varphi(x)}}{\partial{u}}, &x \in P_i\\
\end{cases}
\end{align}
where
\begin{equation}
    q = i+\frac{1}{2}(\varphi(x)+1).
\end{equation}

From (\ref{d_l}) and (\ref{d_u}), we can conclude that the large outlier points are clipped by $u$ and mainly contribute to the update of $u$, while the small ones are clipped by $l$ and mainly contribute to the update of $l$. Data points falling in the middle range will influence derivative of both $u$ and $l$. When clipping error dominates the whole quantization error, the outliers' gradients will be large and thus serve as the major power for weight updating. Otherwise, when rounding error dominates the error, points in the middle range will affect more in the process of backward propagation.

Now we can see our DSQ function has three key parameters: $\alpha$, $l$ and $u$, all of which can be optimized during training. By optimizing the clipping values jointly with the similarity factor $\alpha$, we not only find an evolutional and differentiable approximation to the standard quantization function, but also balance the clipping error and rounding error, which together help bridge the accuracy gap between the full-precision and extremely low-bit quantized models.

\subsection{Training and Deploying} \label{method:deploy}
In this paper DSQ function is proposed with the evolution training to optimize both the DSQ and network parameters, aiming to fine-tune a quantized network from full-precision network. The detailed fine-tuning procedures for the convolution network are listed in Algorithm \ref{algo}.
\vspace{-0.08in}
\begin{algorithm}
	\caption{Feed-Forward and Back-Propagation procedures for a convolution layer quantized with DSQ.} 
	\label{algo}
	\begin{algorithmic}[1]
		\STATEx \textbf{Input:} the input activation $\mathbf a$
		\STATEx \textbf{Parameters:} weight $\mathbf w$, clip value $\mathbf{l_a}$, $\mathbf{u_a}$, $\mathbf{l_w}$, $\mathbf{u_w}$ and similarity factor $\mathbf{\alpha_a}$, $\mathbf{\alpha_w}$
		\STATEx \textbf{Output:} the output activation $\mathbf o$

		\STATEx {\textbf{Feed-Forward}}
		\STATE \quad Clip $\mathbf a$ with $\mathbf{l_a}$,$\mathbf{u_a}$ and clip $\mathbf w$ with $\mathbf {l_w}$, $\mathbf{u_w}$
		\STATE \quad Apply asymptotic function $\varphi$ to activation and weight
		\STATEx \quad \quad$\mathbf{a_{sq}} = \varphi_{a}(\mathbf a)$
		\STATEx \quad \quad $\mathbf{w_{sq}} = \varphi_{w}(\mathbf w)$ 
		\STATE \quad Keep consistent with standard quantization
		\STATEx \quad \quad $\mathbf{a_{q}} = \text{sgn}(\mathbf{a_{sq}})$
		\STATEx \quad \quad $\mathbf{w_{q}} = \text{sgn}(\mathbf{w_{sq}})$
		\STATE \quad Dequantize $\mathbf{a_{q}}$ and $\mathbf{w_{q}}$
		\STATEx \quad \quad  $\mathbf{\hat{a}}=\mathbf{l_a}+\Delta_a(\mathbf{i}+\frac{\mathbf{a_{q}}+1}{2})$
		\STATEx \quad \quad  $\mathbf{\hat{w}}=\mathbf{l_w}+\Delta_w(\mathbf{j}+\frac{\mathbf{w_{q}}+1}{2})$
		\STATE \quad Calculate the output: $\mathbf{o}=\textit{Convolution}(\mathbf{\hat{w}},  \mathbf{\hat{a}})$
		\STATEx {\textbf{Backward-Propagation}}
		\STATE \quad Calculate the gradients(take $\mathbf{a}$ as an example)
		\STATEx \quad \quad  $\frac{\partial{\mathcal{L}}}{\partial{\mathbf a}}=\frac{\partial{\mathcal{L}}}{\partial \mathbf{\hat{a}}} \frac{\partial{\mathbf{\hat{a}}}}{\partial{\mathbf{a_q}}} \frac{\partial{\mathbf{a_{sq}}}}{\partial{\mathbf{a}}}$

		\STATEx \quad \quad  $\frac{\partial{\mathcal{L}}}{\partial{\mathbf {\alpha_a}}}=\frac{\partial{\mathcal{L}}}{\partial \mathbf{\hat{a}}} \frac{\partial{\mathbf{\hat{a}}}}{\partial{\mathbf{a_q}}} \frac{\partial{\mathbf{a_{sq}}}}{\partial{\mathbf{\alpha_a}}}$
		
		\STATEx \quad \quad  
		$\frac{\partial{\mathcal{L}}}{\partial{\mathbf {l_a}}}=\frac{\partial{\mathcal{L}}}{\partial \mathbf{\hat{a}}} \{1+(\mathbf{i}+\frac{\mathbf{a_{q}}+1}{2}) \frac{\partial{\Delta_a}}{\partial{\mathbf{l_a}}}+\frac{\Delta_a}{2} \frac{\partial\mathbf{a_{sq}}}{\partial\mathbf{l_a}}\}$
		
		\STATEx \quad \quad  $\frac{\partial{\mathcal{L}}}{\partial{\mathbf {u_a}}}=\frac{\partial{\mathcal{L}}}{\partial \mathbf{\hat{a}}} \{(\mathbf{i}+\frac{\mathbf{a_{q}}+1}{2})\frac{\partial{\Delta_a}}{\partial{\mathbf{u_a}}}+\frac{\Delta_a}{2} \frac{\partial\mathbf{a_{sq}}}{\partial\mathbf{u_a}}\}$

		\STATEx {\textbf{Parameters Update}}
		\STATE \quad Update all parameters with the learning rate $\eta$
	\end{algorithmic}
\end{algorithm}
\vspace{-0.1in}

\begin{figure}[htbp]
    \vspace{-0.15in}
	\begin{center}
		\includegraphics[width=1\linewidth]{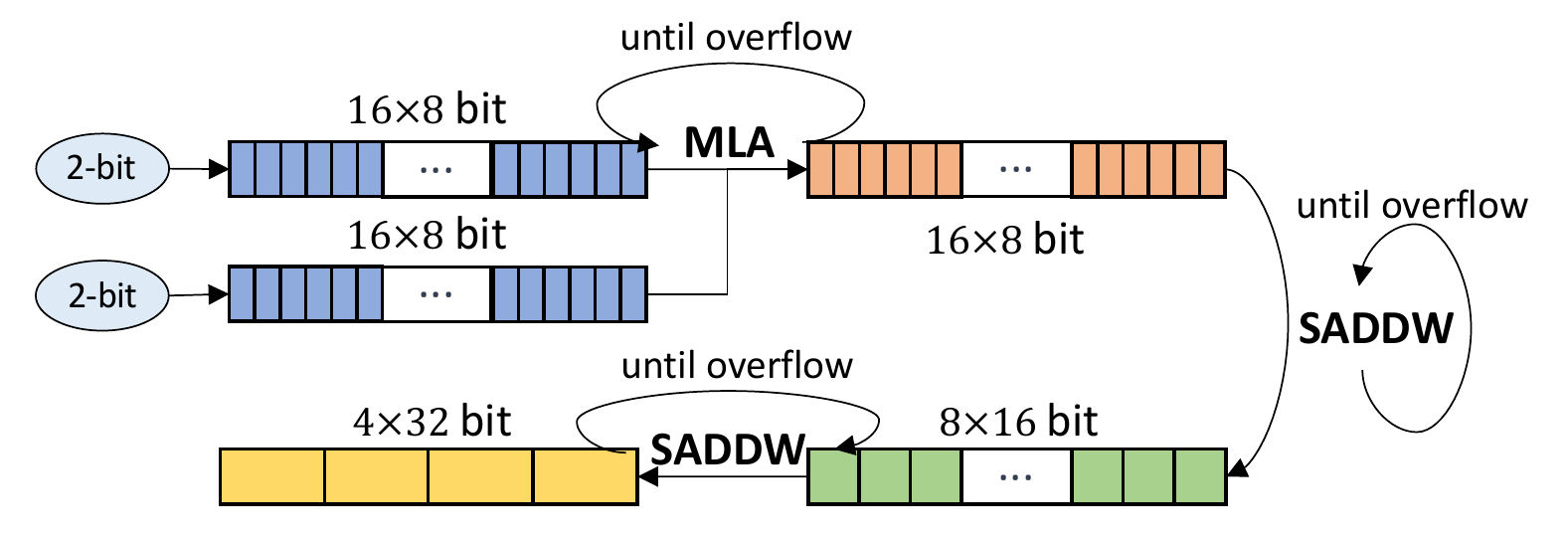}
	\end{center}
	\vspace{-0.2in}
	\caption{Data flow in our fast GEMM implementation.}
	\label{fig:data_flow}
	\vspace{-0.1in}
\end{figure}

For deploying on devices with limited computing resources, we also implement the low-bit computation kernels to accelerate the inference on ARM architecture. In the convolution networks, multiply and accumulation are the core operations of General Matrix Multiply (GEMM), which can be efficiently completed by the MLA instruction on ARM NEON. MLA multiplies two numbers in 8-bit registers and accumulates the result into an 8-bit register. In case that the accumulator is nearly overflowed, we can transfer the value to a 16-bit register by the SADDW instruction. Figure \ref{fig:data_flow} shows the complete data flow of our GEMM kernel.

Though SADDW takes extra computational cost than MLA, fortunately we can reduce the probability that the data transferring occurs.  Given two $b$-bit signed numbers, we can call MLA instruction up to $\frac{2^7-1}{(-2^{b-1})^2}$ times until transferring the result to the 16-bit register by SADDW. The ratio of times for calling MLA and 16-bit register SADDW is listed in Table \ref{instruction_ratio}. It can be concluded that better acceleration performance will be achieved with lower quantized bits. In practice, our low-bit GEMM kernels can outperform the other open-source inference frameworks.

\begin{table}[htbp]
    \vspace{-0.1in}
	\caption{Ratio of time for calling MLA (8-bit register) and SADDW (16-bit register) with respect to different number of quantized bits.}
	\label{instruction_ratio}
	\centering
	\begin{tabular}{c|c|c|c}
		\hline
		$b$ & 2 & 3 & 4 \\
		\hline
        MLA/SADDW & 31/1 & 7/1 & 1/1 \\
        \hline
	\end{tabular}
	\vspace{-0.27in}
\end{table}

\section{Experiments}
In this section, we conduct extensive experiments to demonstrate the effectiveness of the proposed DSQ, on two popular image classification datasets: CIFAR-10 \cite{CIFAR} and ImageNet (ILSVRC12) \cite{Imagenet}. The CIFAR-10 dataset consists of 50K training images and 10K testing images of size 32$\times$32 with 10 classes. ImageNet ILSVRC12 contains about 1.2 million training images and 50K testing images with 1,000 classes.

\subsection{Settings}

\noindent\textbf{DSQ function:} We implement DSQ using Pytorch \cite{pytorch}, as a flexible module that can be easily inserted to the binary or uniform quantization models. Because the DSQ function is differentiable, it can be implemented directly with Pytorch's automatic differentiation mechanism. There are two ways to quantize model to 1-bit, i.e., the binarization with $\{-1, +1\}$ and the uniform quantization. Our DSQ function is compatible with both approaches. When building up a quantized model, we simply insert DSQ function to all places that will be quantized, e.g., the inputs and weights of a convolution layer.

\noindent\textbf{Network structures:}
We employ the widely-used network structures including VGG-Small \cite{LQNet}, ResNet-20 for CIFAR-10, and ResNet-18, ResNet-34 \cite{Resnet}, MobileNetV2 \cite{mobilenetv2} for ImageNet. For binarized models, we adopt parameter-free type-A shortcut as \cite{PreactResnet} and apply the activation function replacement introduced by \cite{BNN}. All convolution and fully-connected layers except the first and the last one are quantized with DSQ.

\noindent\textbf{Initialization:} For initialization, we try fine-tuning from a pre-train model and training from scratch. For hyper-parameters, we follow the rules described in the origin papers \cite{PACT, Resnet, PreactResnet, LQNet}. For parameter $\alpha$, we choose 0.2 as an initial value. For clipping value $l$ and $u$, we try the following two strategies: moving average statistics and optimization by backward propagation.

\subsection{Analysis of DSQ}
First, we empirically analyze DSQ from the aspects of rectification, convergence, evolution, etc.
\vspace{-0.12in}
\subsubsection{Rectification}
\begin{figure}[htbp]
\vspace{-0.08in}
\centering
\begin{minipage}[t]{1\linewidth}
\centering
layer2.2 conv2 filter\#1
\vspace{-0.1in}
\end{minipage}
\begin{minipage}[t]{0.30\linewidth}
\centering
\includegraphics[width=1\linewidth]{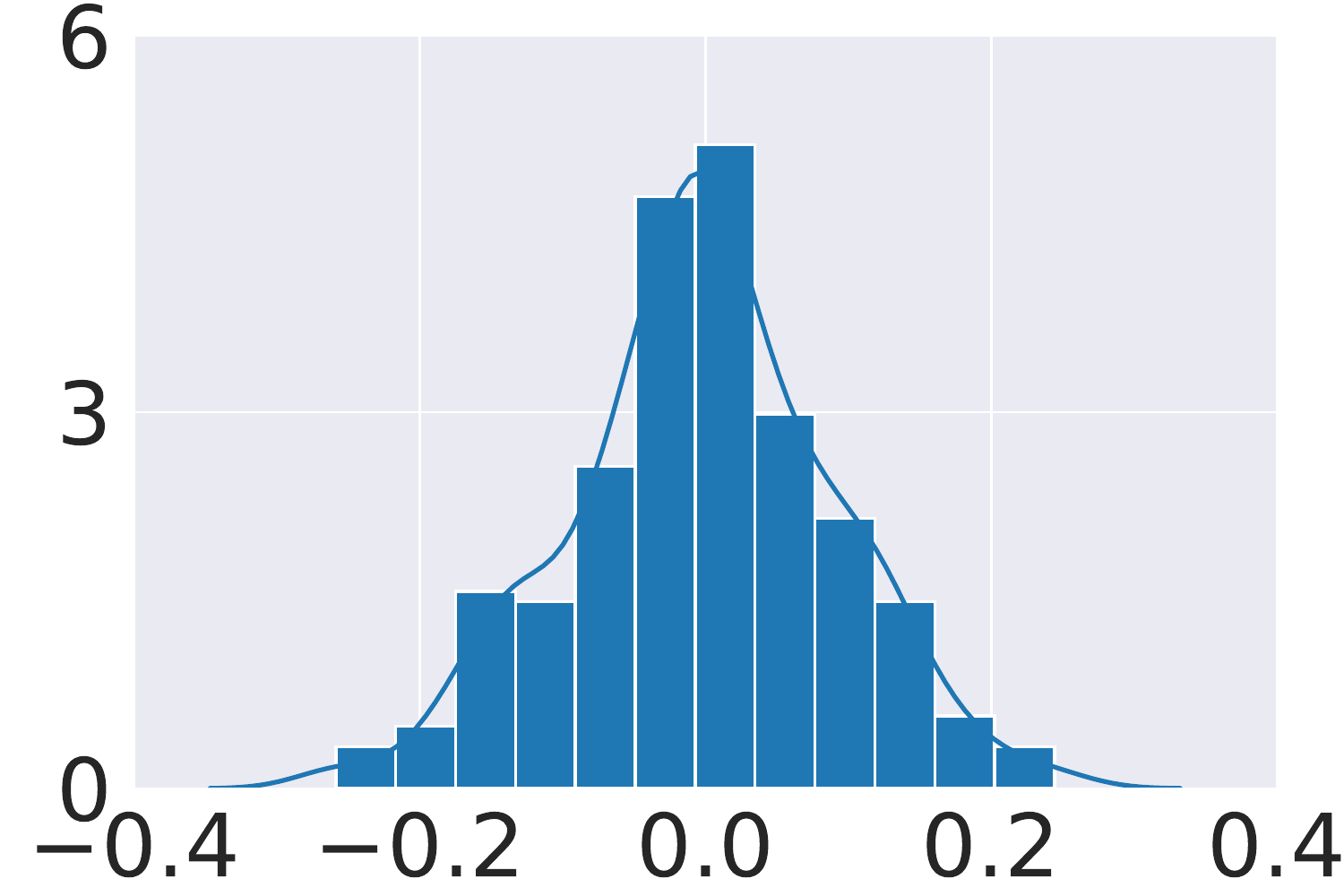}
\end{minipage}
\hfill
\begin{minipage}[t]{0.30\linewidth}
\centering
\includegraphics[width=1\linewidth]{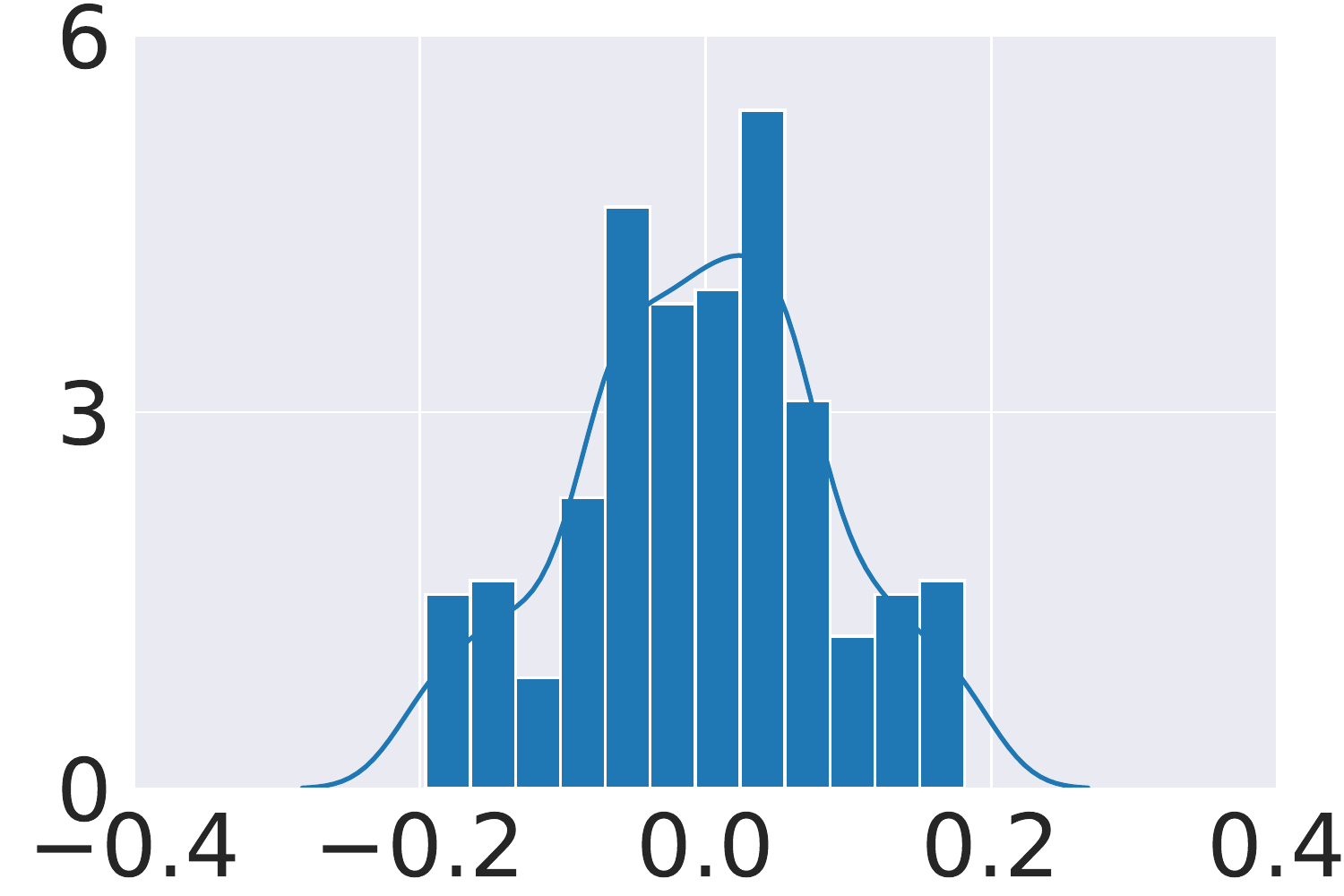}
\end{minipage}
\hfill
\begin{minipage}[t]{0.30\linewidth}
\centering
\includegraphics[width=1\linewidth]{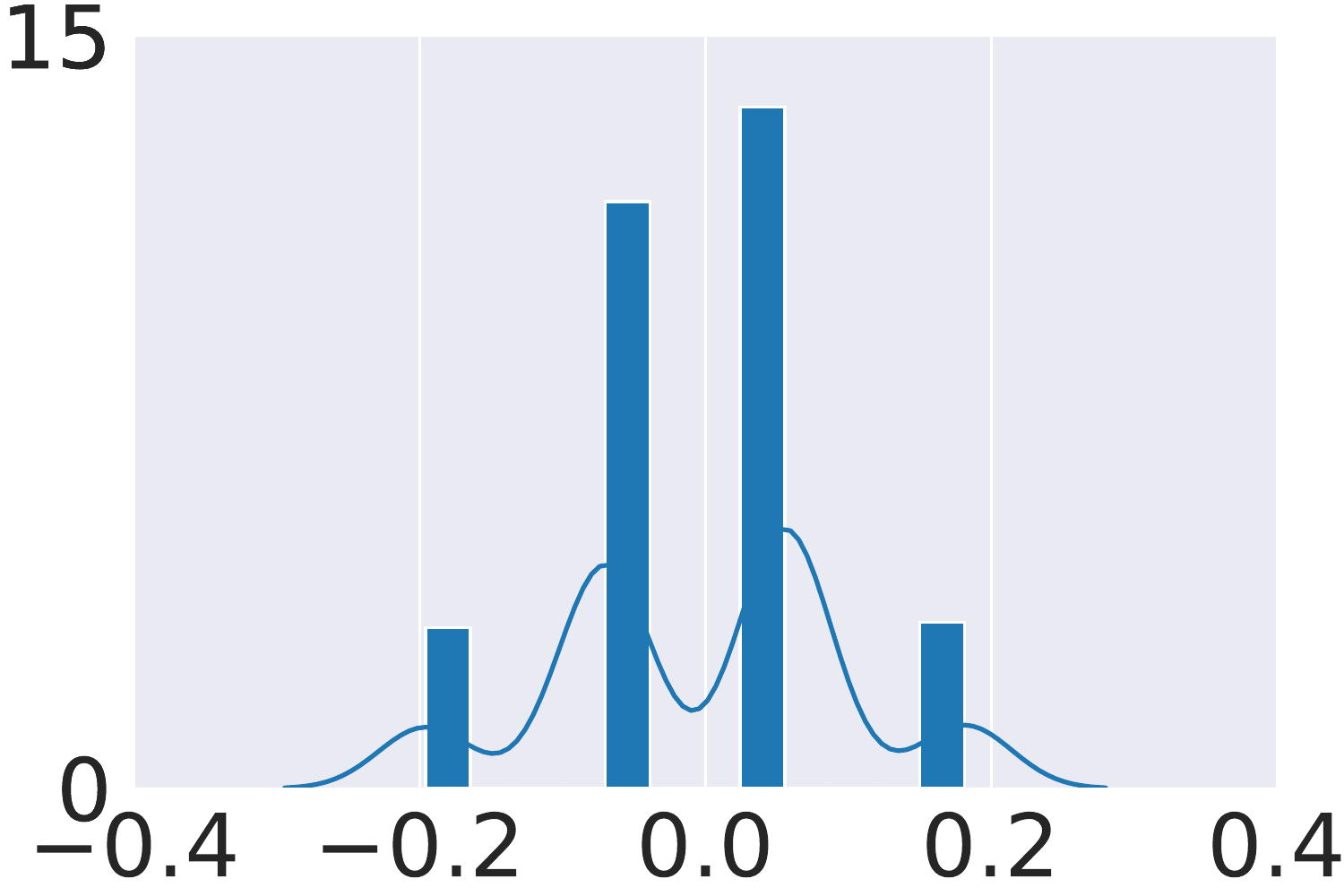}
\end{minipage}
\vspace{-0in}
\begin{minipage}[t][][b]{1\linewidth}
\centering
layer3.2 conv2 filter\#1
\vspace{-0.1in}
\end{minipage}
{\subfigcapskip = -4pt
\subfigure[Before DSQ]{
\begin{minipage}[t]{0.30\linewidth}
\centering
\includegraphics[width=1\linewidth]{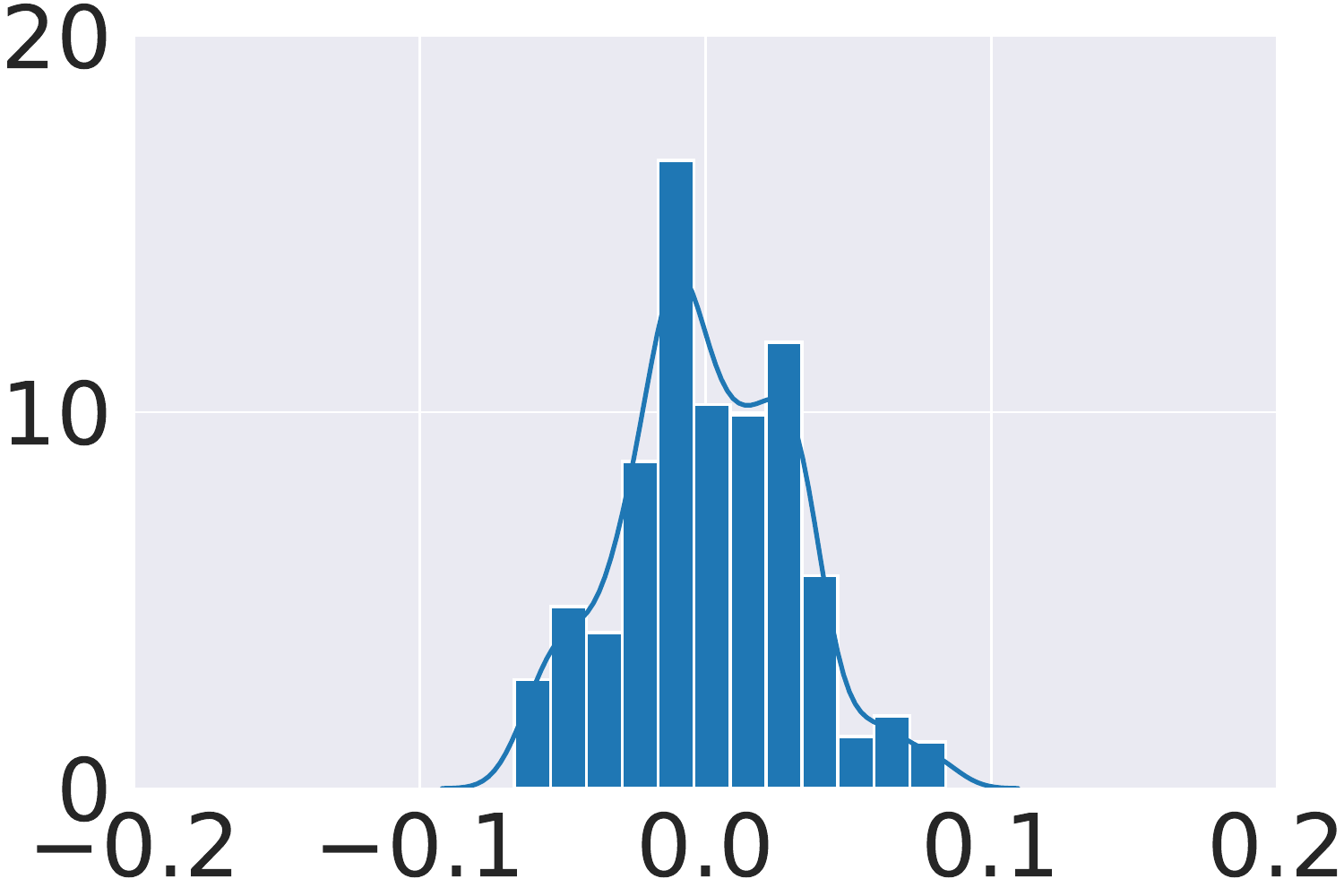}
\end{minipage}
}}
\hfill
{\subfigcapskip = -4pt
\subfigure[After DSQ]{
\begin{minipage}[t]{0.30\linewidth}
\centering
\includegraphics[width=1\linewidth]{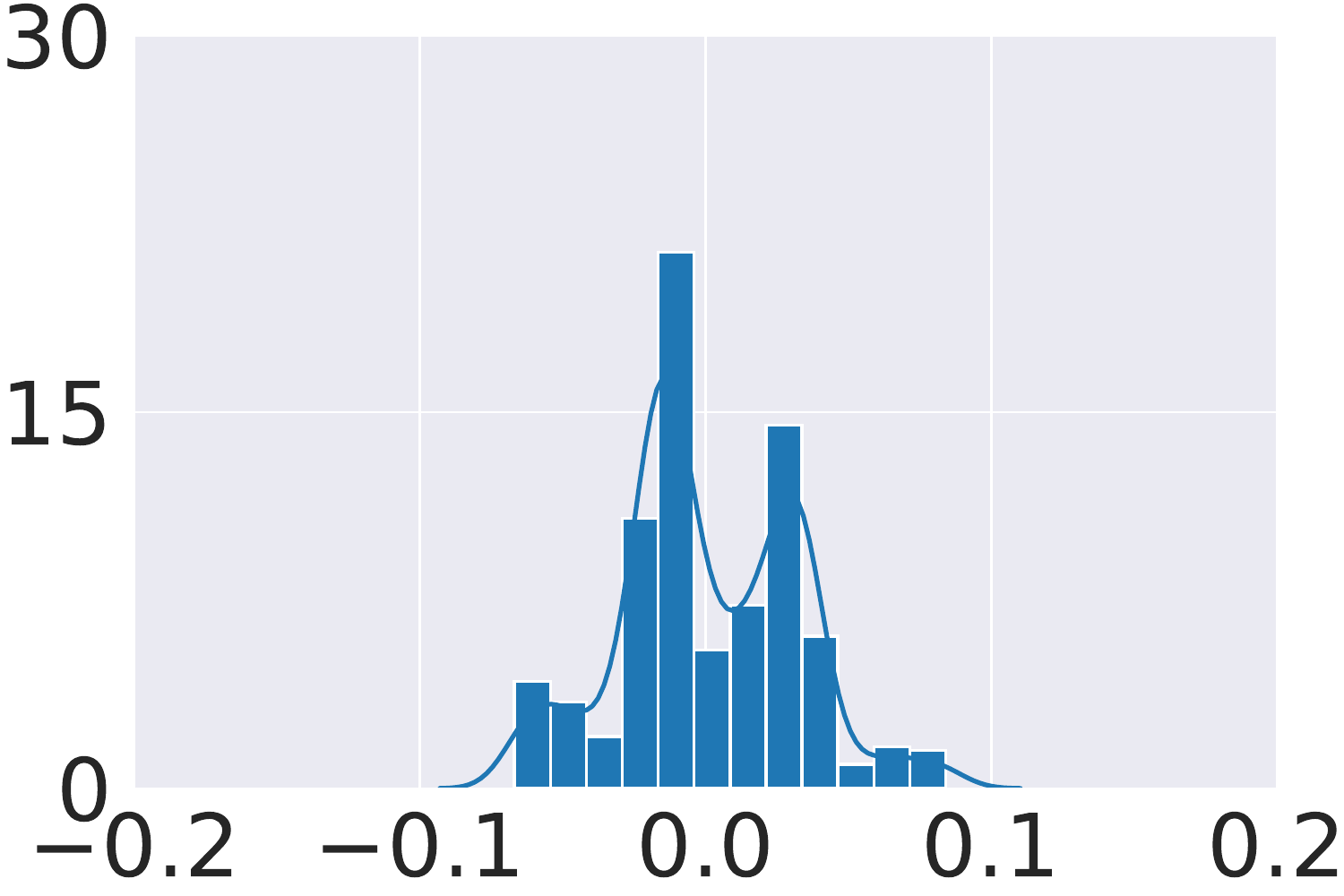}
\end{minipage}
}}
\hfill
{\subfigcapskip = -4pt
\subfigure[After sign]{
\begin{minipage}[t]{0.30\linewidth}
\centering
\includegraphics[width=1\linewidth]{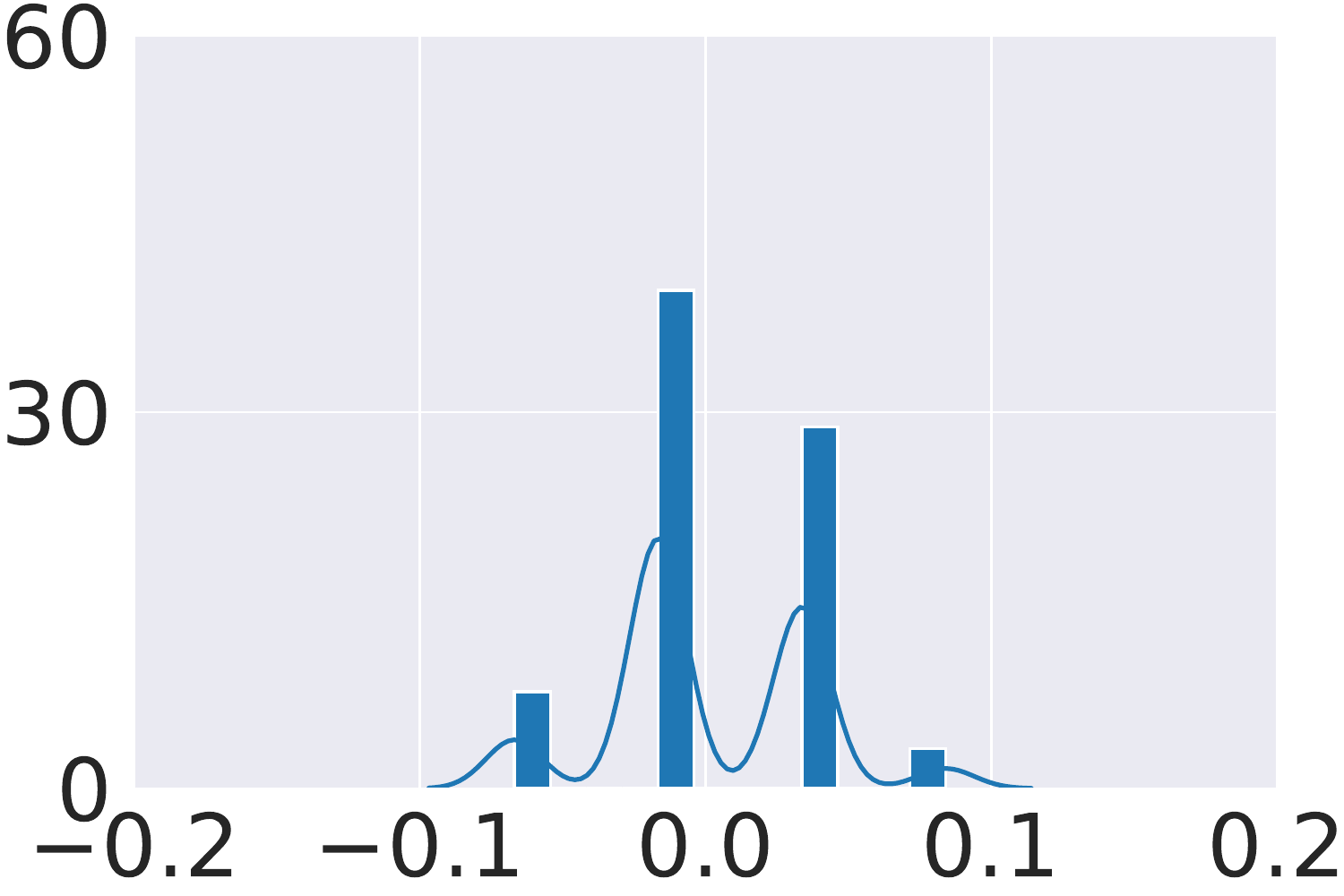}
\end{minipage}
}}
\caption{Data distribution before and after DSQ (2-bit).}
\label{fig:redistribution}
\vspace{-0.28in}
\end{figure}

An important effect of DSQ function is to redistribute the data and align them to the quantization values, which subsequently decrease the backward propagation error. To investigate this point, in Figure \ref{fig:redistribution} we visualize the weight's distribution of ResNet-20 on CIFAR-10 before and after 2-bit DSQ function. 
From the figure, we can observe that the data distribution, originally similar to normal distribution (Figure \ref{fig:redistribution}(a)), appears with a few peaks in the histogram after DSQ's rectification (Figure \ref{fig:redistribution}(b)), and subsequently the data can be completely quantized to 4 quantization points after the sign function (Figure \ref{fig:redistribution}(c)). This observation proves that DSQ serves as a promising bridge between the origin full-precision model and low-bit quantized model, largely reducing the quantization loss in practice.

\subsubsection{Convergence}\label{converge}
State-of-the-art binary/uniform quantized networks often adopt the training strategy that directly applies quantization in forwarding process, but STE in backward process. They ignore the negative effect of the quantization loss on the gradient calculation, and thus often face the unstable training in most cases. We here show that training these models with our DSQ can significantly improve the ability to converge. 

\begin{figure}[htbp]
\vspace{-0.18in}
\centering
{\subfigcapskip = -4pt
\subfigure[VGG-Small]{
\begin{minipage}[t]{0.45\linewidth}
\centering
\includegraphics[width=1\linewidth]{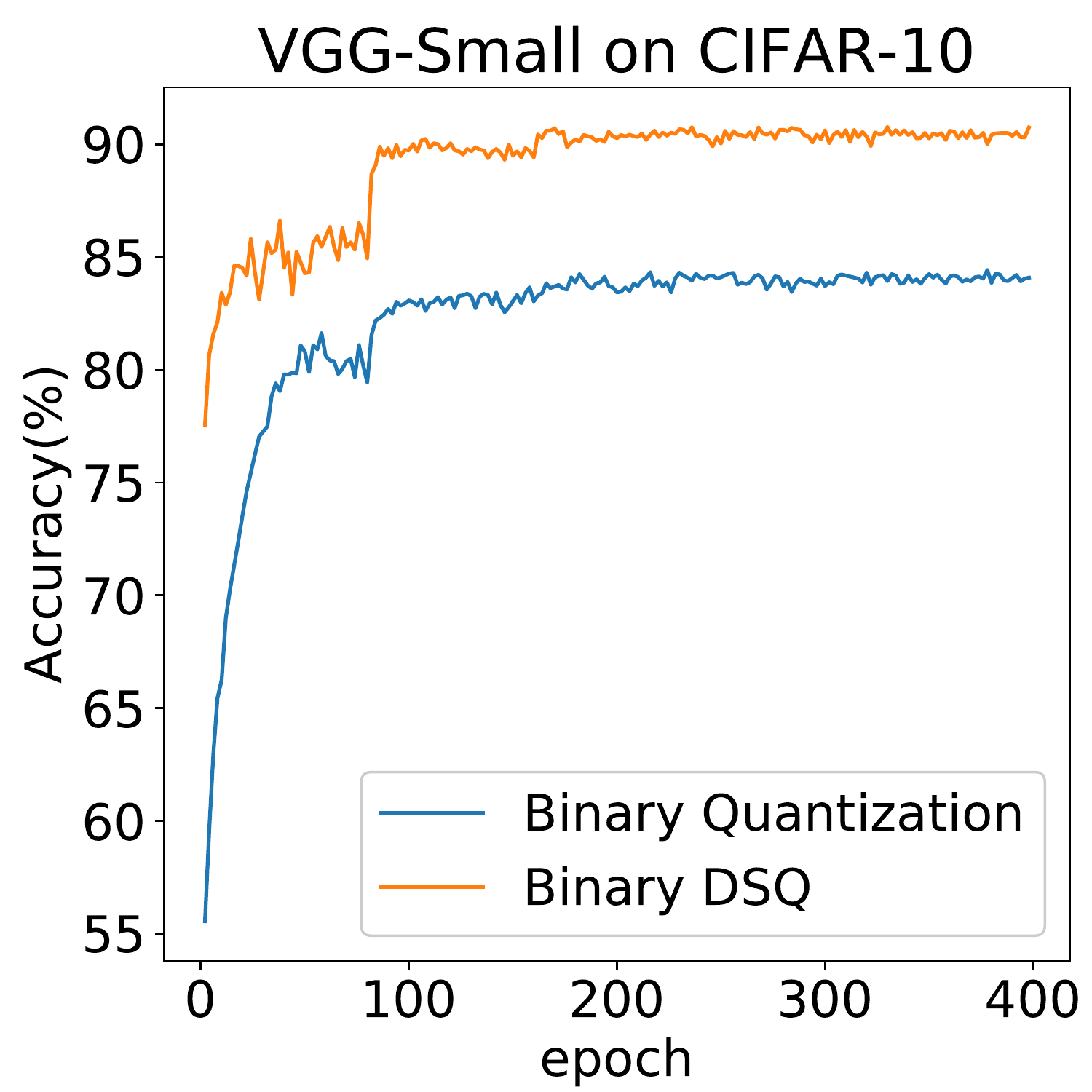}
\end{minipage}
}}
{\subfigcapskip = -4pt
\subfigure[ResNet-34]{
\begin{minipage}[t]{0.45\linewidth}
\centering
\includegraphics[width=1\linewidth]{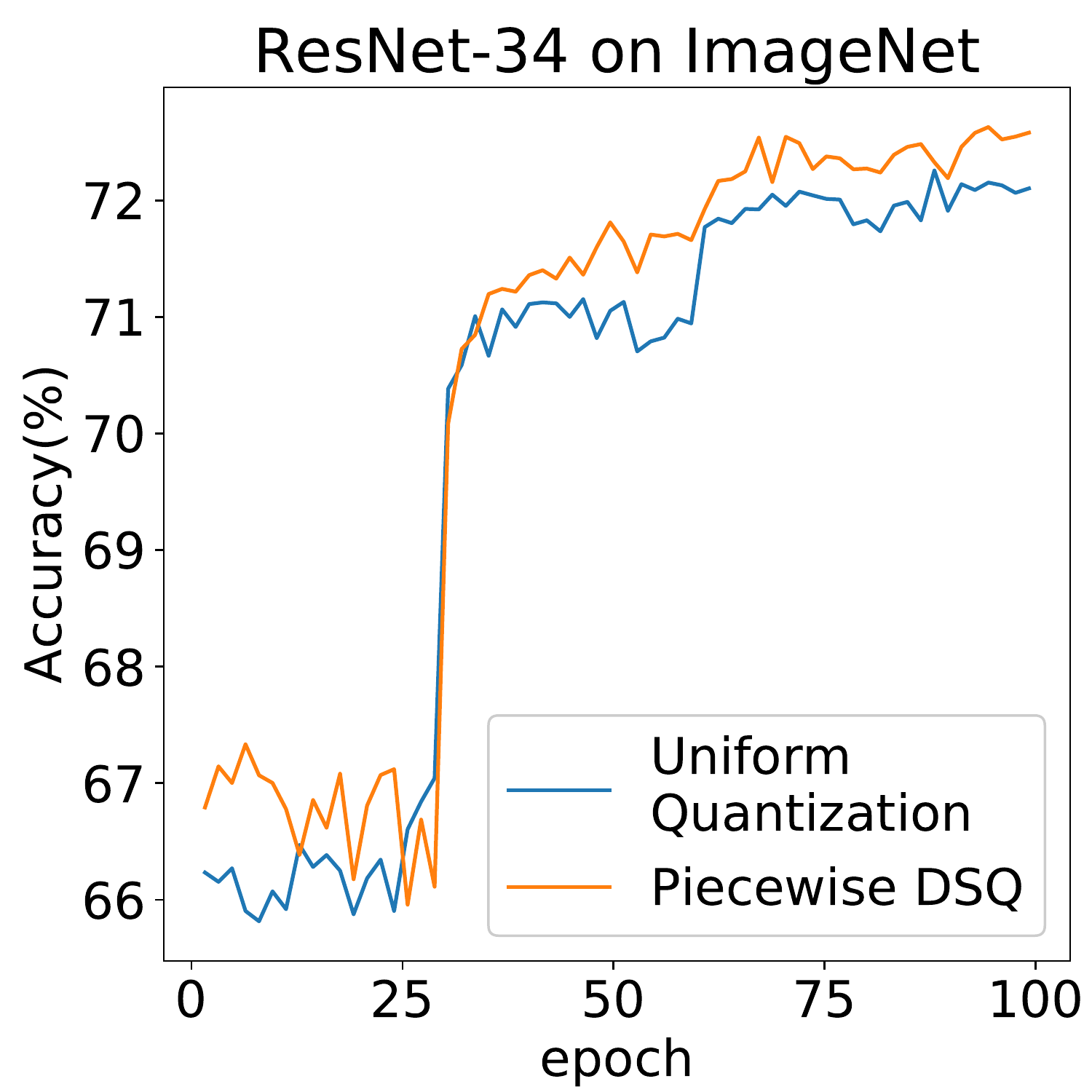}
\end{minipage}
}}
\caption{Training quantized model with/without DSQ function.}
\label{fig:converge}
\vspace{-0.28in}
\end{figure}

With the redistribution of DSQ, the numerical difference between quantized data and full-precision data is decreased, contributing to a more accurate backward. From another perspective, the introducing of DSQ can be viewed as an optimizer of STE, which can improve the ability to converge in process of optimization. Figure \ref{fig:converge} compares the accuracy curves of validation with and without DSQ when using binarization in VGG-Small on CIFAR-10 and 3-bit uniform quantization in ResNet-34 on ImageNet. We can find that training with DSQ can achieve higher accuracy.

\subsubsection{Evolution}
\begin{figure}[htbp]
\vspace{-0.2in}
\centering
{\subfigcapskip = -4pt
\subfigure[$\alpha$ for activations]{
\begin{minipage}[t]{0.45\linewidth}
\centering
\includegraphics[width=1\linewidth]{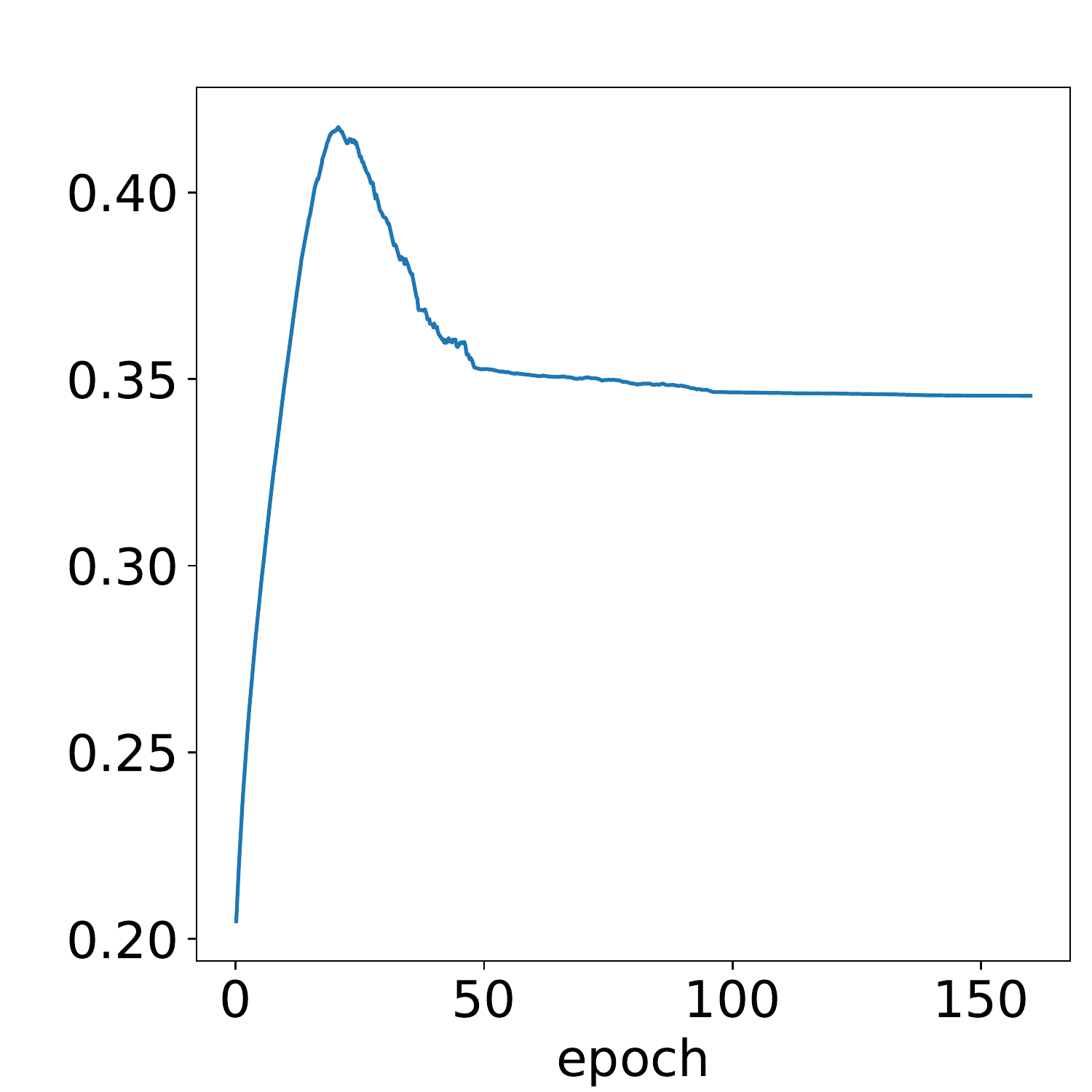}
\end{minipage}
}}
{\subfigcapskip = -4pt
\subfigure[$\alpha$ for weights]{
\begin{minipage}[t]{0.45\linewidth}
\centering
\includegraphics[width=1\linewidth]{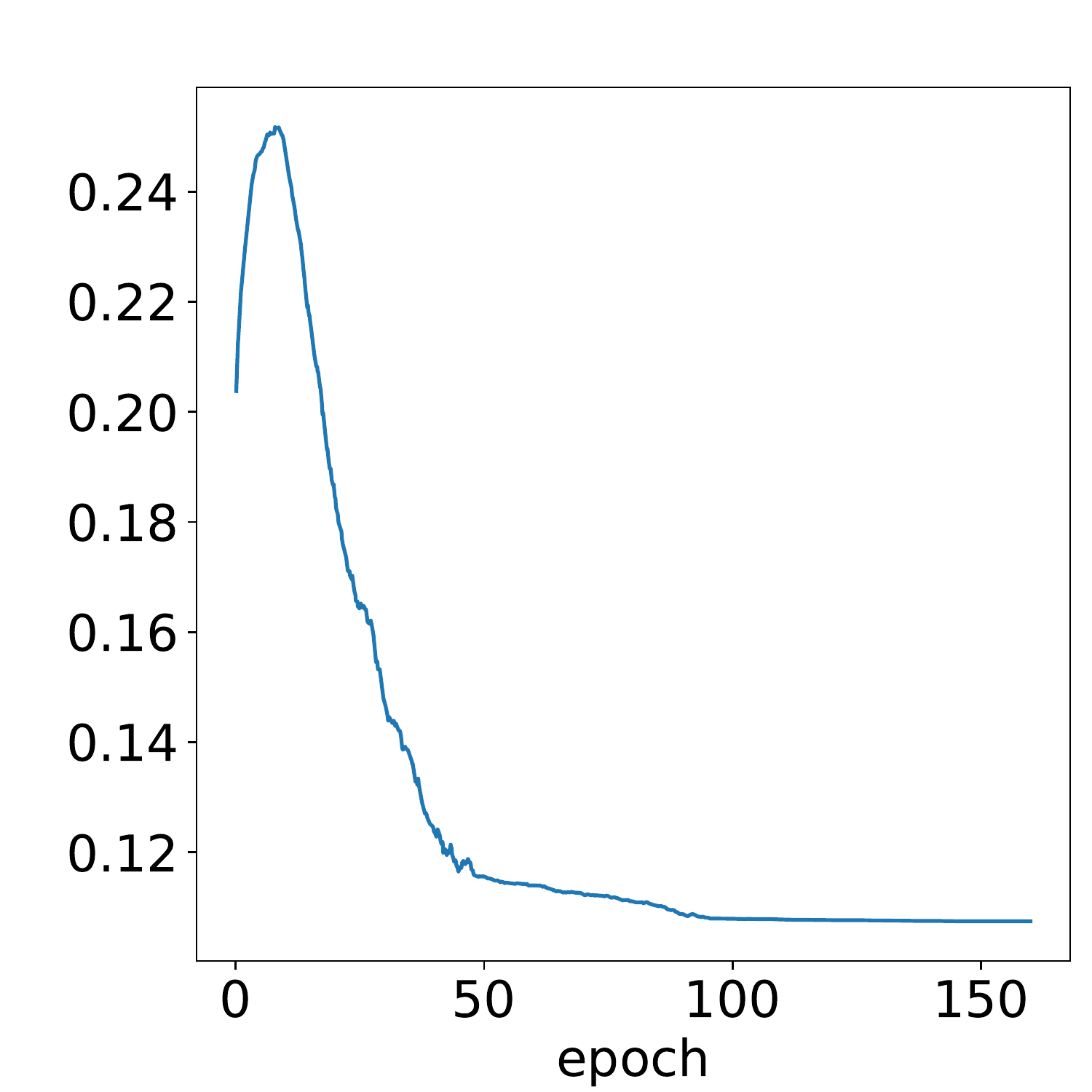}
\end{minipage}
}}
\caption{Automatic evolution of $\alpha$ during training.}
\label{fig:evolution}
\vspace{-0.08in}
\end{figure}

The automatic evolution of DSQ function is the key to the feasible approximation to the standard quantization, which is determined by the similarity factor $\alpha$. In our experiments, for both weights and activations of ResNet-20 on CIFAR-10, $\alpha$ is initialized to 0.2, and to ensure the stability, we limit it to a reasonable range of $(0, 0.5)$ with $k\leq 1000$. Figure \ref{fig:evolution} (a) and (b) respectively plot the curves of $\alpha$ per step for activations and weights during the evolution training.

For both activations and weights, we can see that at the beginning of training, $\alpha$ will increase sharply. Considering when $\alpha$ becomes large, DSQ behaves more like an identity operation, this phenomenon implies that we should not quantize much at the beginning. After that, $\alpha$ decreases gradually and finally converges to a stable value, which makes DSQ approximate the standard quantization. In the whole training process, $\alpha$ is automatically adjusted according to the network loss, and thus enables the stronger flexibility of DSQ than the incremental quantization relying on manually adjusting.

\begin{table}[htbp]
	\centering
		\caption{Learnt $\alpha$ for activations and weights of ResNet-20}
	\label{alpha_per_layer}
	\begin{tabular}{ccc}
		\hline
		Layer & Weight & Activation \\
		\hline
layer1.0.conv1 & 0.1075 & 0.3455 \\
layer1.0.conv2 & 0.0950 & 0.3101 \\
layer1.1.conv1 & 0.0895 & 0.3046 \\
layer1.1.conv2 & 0.0868 & 0.2526 \\
layer1.2.conv1 & 0.1368 & 0.3621 \\
layer1.2.conv2 & 0.0926 & 0.3401 \\
layer2.0.conv1 & 0.1578 & 0.3903 \\
layer2.0.conv2 & 0.1810 & 0.3785 \\
layer2.0.downsample.0 & \textbf{0.0828} & 0.2887 \\
layer2.1.conv1 & 0.1641 & 0.2722 \\
layer2.1.conv2 & 0.1162 & 0.2663 \\
layer2.2.conv1 & 0.1605 & 0.2690 \\
layer2.2.conv2 & 0.1059 & 0.2440 \\
layer3.0.conv1 & 0.2042 & 0.3993 \\
layer3.0.conv2 & \textbf{0.2779} & \textbf{0.4532} \\
layer3.0.downsample.0 & 0.0914 & \textbf{0.2327} \\
layer3.1.conv1 & 0.2484 & 0.4241 \\
layer3.1.conv2 & 0.2301 & 0.3918 \\
layer3.2.conv1 & 0.1965 & 0.4238 \\
layer3.2.conv2 & 0.0975 & 0.3277 \\
		\hline
	\end{tabular}
	\vspace{-0.08in}
\end{table}

Table \ref{alpha_per_layer} also reports the final optimal $\alpha$ of each layer. First, we find that usually $\alpha$ of weights is smaller than that of activations. This means that in the low-bit quantized network usually weights are more tolerant to the quantization, while activations are more sensitive, which conforms to the experiences and conclusions of previous studies \cite{Dorefa,PACT}. Second, different layers show different sensitivity to the quantization. For example, the downsampling convolution layers can be quantized much (a small $\alpha$), while some layers such as layer3.0.conv2 are not suitable for quantization (a large $\alpha$). This conclusion is very useful for understanding and improving the network quantization.

\subsection{Ablation study}
To further understand how DSQ works in practice, we also conduct the ablation study on both model binarization and uniform quantization. We specify $\alpha$ as 0.2 for network binarization experiments. From Table \ref{binary_ablation}, we can see that even the naive DSQ with the fixed $\alpha$ brings stable improvement over the basic binarization strategy that directly applies sign function over ResNet-20 on CIFAR-10, e.g., nearly 2\% performance gain for the 1W1A (1-bit quantization for both weight and activation) case.

\begin{table}[htbp]
	\centering
	\caption{Ablation study on 1-bit binarized quantization.}
	\label{binary_ablation}
	\begin{tabular}{ccc}
		\hline
		Method & Bit-Width (W/A)&Accuracy (\%)\\\hline
		FP&32/32&90.84\\\hline
		Binary&1/1&82.46\\
		Binary DSQ&1/1&83.80\\
		Piecewise DSQ & 1/1 & \textbf{84.11} \\\hline
		Binary&1/32&90.11\\
		Binary DSQ & 1/32 & \textbf{90.24}\\
		Piecewise DSQ & 1/32& 90.03\\
		\hline
	\end{tabular}
	\vspace{-0.18in}
\end{table}

In Table \ref{uniform_ablation}, we further study the effect of evolution training (learnt $\alpha$) and balanced quantization error (learnt $l, u$) over the 2-bit uniform quantization of ResNet-20 on CIFAR-10. It is obvious that learning the adaptive $\alpha$ and the clipping values $l, u$ respectively brings accuracy improvement. Besides, since DSQ can be conveniently inserted to any standard quantization method, we further investigate its performance over the state-of-the-art  quantization method PACT \cite{PACT}. We implement the PACT method and the results in Table \ref{uniform_ablation} show that DSQ can further boost the performance of PACT, proving the flexibility and generality of DSQ.

\begin{table}[htbp]
    \vspace{-0.08in}
	\centering
	\caption{Ablation study on 2-bit uniform quantization.}
	\label{uniform_ablation}
	\begin{tabular}{cccc}
		\hline
		Method&Top-1 (\%)&Top-5 (\%)\\		\hline
		Standard Quantization & 86.63 & 99.35\\
		Fixed $\alpha$ & {86.95} & 99.50\\
		Learnt $\alpha$ & {87.25} & 99.49\\
		Learnt $\alpha, l, u$ & \textbf{88.44} & \textbf{99.50}\\
		\hline
	\end{tabular}
	
\end{table}

\begin{table}[htbp]
    \vspace{-0.2in}
	\centering
    \caption{Ablation study on 2-bit PACT.}
	\label{pact_ablation}
	\begin{tabular}{ccc}
		\hline
		Method&Top-1 (\%)&Top-5 (\%)\\
		\hline
		PACT&88.24&99.60\\
		PACT+DSQ & \textbf{90.11} & \textbf{99.71}\\
		\hline
	\end{tabular}
    \vspace{-0.18in}
\end{table}

\subsection{Comparison with State-of-the-arts}
Next we comprehensively evaluate DSQ by comparing it with the existing state-of-the-art quantization methods.

\textbf{Comparison on CIFAR-10:} Table \ref{cifar10} lists the performance using different methods on CIFAR-10, respectively including BNN \cite{BNN}, XNOR-Net \cite{XnorNet} over VGG-Small, and DoReFa-Net \cite{Dorefa}, LQ-Net \cite{LQNet} over ResNet-20. All methods quantize weights or activations to 1 bit. In all cases, our DSQ obtains the best performance for the two different network structures. More importantly, when using 1-bit activations and 1-bit weights (1/1), our method gets very significant improvement (i.e., 84.11\% v.s. 79.30\%) over the state-of-the-art DoReFa-Net \cite{Dorefa}. Note that for VGG-Small, DSQ using 1-bit weights and activations can even obtain better performance than the full-precision model.

\begin{table}[htbp]
	\centering
	\caption{Comparison of 1-bit quantized models on CIFAR-10.}
	\label{cifar10}

    \begin{tabular}{cccc}
        \hline
        Model & Method & \tabincell{c}{Bit-Width \\(W/A)} & Accuracy (\%) \\
        \hline
        \multirow{4}{*}{VGG-Small}
            & FP & 32/32 & 91.65 \\
            \cline{2-4}
            & BNN \cite{BNN} & 1/1 & 89.90 \\
            & XNOR \cite{XnorNet} & 1/1 & 89.80 \\
            \cline{2-4}
            & Ours & 1/1 & \textbf{91.72} \\
        \hline
        \multirow{6}{*}{ResNet-20}
            & FP & 32/32 & 90.78 \\
            \cline{2-4}
            & DoReFa \cite{Dorefa} & 1/1 & 79.30 \\
            \cline{2-4}
            & Ours & 1/1 & \textbf{84.11} \\
            \cline{2-4}
            & DoReFa \cite{Dorefa} & 1/32 & 90.00 \\
            & LQ-Net \cite{LQNet} & 1/32 & 90.10 \\
            \cline{2-4}
            & Ours & 1/32 & \textbf{90.24} \\
        \hline
    \end{tabular}
\end{table}

\begin{table}[htbp]
    \vspace{-0.05in}
	\centering
		\caption{Comparison of different quantized models on ImageNet.}
	\label{imagenet}
	\begin{threeparttable}
	\begin{tabular}{cccc}
		\hline
		Model & Method & \tabincell{c}{Bit-Width\\(W/A)} & Accuracy (\%) \\
        \hline
		\multirow{12}{*}{ResNet-18}
    		& FP&32/32 & 69.90 \\
    		\cline{2-4}
    		& BWN \cite{XnorNet} & 1/32 & 60.80 \\
    		& HWGQ \cite{HWGQ} & 1/32 & 61.30 \\
    		& TWN \cite{TWN} & 2/32 & 61.80 \\
            \cline{2-4}
    		& Ours & 1/32 & \textbf{63.71} \\
    		\cline{2-4}
    		& PACT \cite{PACT} & 2/2 & 64.40 \\
    		& LQ-Net \cite{LQNet} & 2/2 & 64.90 \\
    		\cline{2-4}
    		& Ours & 2/2 & \textbf{65.17} \\
    		\cline{2-4}
    		& ABC-Net \cite{ABCNet} & 3/3 & 61.00 \\
    		& PACT \cite{PACT} & 3/3 & 68.10 \\
    		& LQ-Net \cite{LQNet} & 3/3 & 68.20 \\
    		\cline{2-4}
    		& Ours & 3/3 & \textbf{68.66} \\
    		\cline{2-4}
    		& BCGD \cite{BCGD}&4/4&67.36\textsuperscript{\textdagger}\\
    		\cline{2-4}
    		& Ours&4/4&\textbf{69.56\textsuperscript{\textdagger}}\\
		\hline
		\multirow{6}{*}{ResNet-34}
    		& FP & 32/32 & 73.80 \\
    		\cline{2-4}
    		& LQ-Net \cite{LQNet}&2/2&69.80\\
    		\cline{2-4}
    		& Ours&2/2&\textbf{70.02}\\
    		\cline{2-4}
    		& ABC-Net \cite{ABCNet}&3/3&66.70\\
    		& LQ-Net \cite{LQNet}&3/3&71.90\\
    		\cline{2-4}
    		& Ours&3/3&\textbf{72.54}\\
    		\cline{2-4}
    		& BCGD \cite{BCGD}&4/4&70.81\textsuperscript{\textdagger}\\
    		\cline{2-4}
    		& Ours&4/4& \textbf{72.76\textsuperscript{\textdagger}}\\
	    \hline
	    \multirow{3}{*}{\tabincell{c}{Mobile-\\NetV2}}
    		& FP & 32/32 & 71.87 \\
    		\cline{2-4}
    		& PACT \cite{PACT,HAQ}&4/4&61.40\\
    		\cline{2-4}
    		& Ours&4/4 &\textbf{64.80}\\
	    \hline
	\end{tabular}
	\begin{tablenotes}
    \footnotesize
    \item[*] The \textdagger\ represents the results of full quantization for activations and weights across all convolution layers.
    \end{tablenotes}
    \end{threeparttable}
    \vspace{-0.25in}
\end{table}

\textbf{Comparison on ImageNet:} For the large-scale case, we study the performance of DSQ over ResNet-18, ResNet-34 and MobileNetV2 on ImageNet. Table \ref{imagenet} shows a number of state-of-the-art quantization methods including BWN \cite{XnorNet}, HWGQ \cite{HWGQ}, TWN \cite{TWN}, PACT \cite{PACT}, LQ-Net \cite{LQNet}, ABC-Net \cite{ABCNet} and BCGD \cite{BCGD}, with respect to different settings. From the table, we can observe that when only quantizing weights over ResNet-18, DSQ using 1 bit outperforms BWN and HWGQ by large margins and even surpasses TWN using 2 bits. Besides, over both ResNet-18 and ResNet-34, the accuracy of our method using 2-bit and 3-bit quantization is also consistenly higher than LQ-Net and ABC-Net. We should point out that LQ-Net is a non-uniform quantization method. This means that our DSQ simultaneously enjoys efficient inference as the uniform methods, and competitive performance to the more complex non-uniform solutions. What's more, the accuracy of DSQ on efficient networks such as MobileNetV2 also significantly exceeds existing methods (e.g., 3.4\% higher than PACT \cite{PACT, HAQ} using 4-bit), which proves the protential of DSQ on hardware-friendly networks with a small amount of parameters.

\subsection{Deploying Efficiency}\label{deploy_speed}
Finally, we highlight the uniqueness of our DSQ that we support extremely low-bit (less than 4-bit) integer arithmetic based on GEMM kernels with ARM NEON 8-bit instructions, while existing open-source high performance inference frameworks (e.g., NCNN-8-bit \cite{ncnn}) usually only support 8-bit operations. In practice, the lower bit width doesn't mean a faster inference speed, mainly due to the overflow and tranferring among the registers as analyzed in Section \ref{method:deploy}. But fortunately, as Table \ref{kernel_self_comparison} shows, our implementation can accelerate the inference even using the extreme lower bits. We also test the real speed of our implementation when quantizing ResNet-18 with DSQ on Raspberry Pi 3B, which has a 1.2 GHz 64-bit quad-core ARM Cortex-A53. As shown in Table \ref{kernel_comparison}, the inference speed using DSQ is much faster than that of NCNN.

\begin{table}[htbp]
    \vspace{-0.05in}
	\centering
    \caption{Time cost (ms) of the typical $3\times 3$ convolution in ResNet using different number of bits (single thread).}\label{kernel_self_comparison}
	\begin{tabular}{ccccc}
    \hline
    input size& \#output & 4-bit & 3-bit & 2-bit\\\hline
    64x56x56& 64 & 43.80 & 40.06 & \textbf{38.11}\\
    128x28x28& 128 & 33.89 & 29.94 & \textbf{28.15}\\
    256x14x14& 256 & 37.03 & 31.16 & \textbf{29.20}\\
    512x7x7& 512 & 30.20 & 26.14 & \textbf{25.43}\\
    \hline
    \end{tabular}
\end{table}

\begin{table}[htbp]
    \vspace{-0.2in}
	\centering
    \caption{Comparison of time cost of ResNet-18 on different inference frameworks with different bits (single thread).}
	\label{kernel_comparison}
	\begin{threeparttable}
	\begin{tabular}{ccc}
        \hline
        & DSQ 2-bit & \href{https://github.com/Tencent/ncnn}{NCNN} 8-bit \cite{ncnn}\\
        \hline
        time (ms) & \textbf{551.22} & 935.51\\
        \hline
	\end{tabular}
	\begin{tablenotes}
    \footnotesize
    \item[*] NCNN was tested with commit d263cd5 on 2019.3.15.
    \end{tablenotes}
    \end{threeparttable}
    \vspace{-0.2in}
\end{table}

\section{Conclusions}
In this paper, we proposed the Differentiable Soft Quantization (DSQ) method to eliminate the accuracy gap between the full-precision networks and low-bit (binary/uniform) quantization networks. DSQ can evolve dynamically during end-to-end training to approximate standard quantization. Since it can reduce both the gradient deviation of backward propagation and the quantization loss in forward inference, state-of-the-art accuracy for various network structures can be promised. As a general module supporting both model binarization and uniform quantization, it also enjoys strong flexibility to improve the performance of different quantization methods, and high  hardware-friendly efficiency based on the fast fix-point GEMM kernels implementation. 

\section*{Acknowledge}
This work was supported by National Natural Science Foundation of China (61690202, 61872021), Fundamental Research Funds for Central Universities (YWF-19-BJ-J-271), Beijing Municipal Science and Technology Commission (Z171100000117022), and State Key Lab of Software Development Environment (SKLSDE-2018ZX-04).

{\small
\bibliographystyle{ieee_fullname}
\bibliography{egbib}
}

\end{document}